\def\year{2019}\relax
\documentclass[letterpaper]{article} %DO NOT CHANGE THIS
\usepackage{aaai19} %Required
\usepackage{times} %Required
\usepackage{helvet} %Required
\usepackage{courier} %Required
\usepackage{url} %Required
\usepackage{graphicx} %Required

\usepackage{color,soul}
\usepackage{amsmath,mathtools}
\usepackage{amsfonts}
\usepackage[capitalize]{cleveref}
\usepackage{algorithm}
\usepackage[noend]{algpseudocode}
\usepackage{multirow,multicol}
\usepackage{MnSymbol}
\usepackage{array}
\usepackage{subcaption}

\DeclareMathOperator*{\ReLU}{\text{ReLU}}
\DeclareMathOperator*{\softmax}{\text{softmax}}
\DeclareMathOperator*{\argmax}{\text{argmax}}
\newcommand\norm[1]{\left\lVert#1\right\rVert}

\newcommand{\citet}[1]
{\citeauthor{#1}~\shortcite{#1}}
\newcommand{\citep}{\cite}

\makeatletter
\def\thanks#1{\protected@xdef\@thanks{\@thanks
        \protect\footnotetext{#1}}}
\makeatother
\begin{document}
\title{DialogueRNN: An Attentive~RNN for Emotion~Detection in Conversations}
\author{Navonil Majumder$^{\dagger\equiv}$, Soujanya Poria$^{\ddagger\equiv}$, Devamanyu Hazarika$^{\Phi}$,\\\textbf{\Large Rada Mihalcea$^{\bigtriangledown}$, Alexander Gelbukh$^{\dagger}$,  Erik Cambria$^{\ddagger}$}\\
$^{\dagger}$Centro de Investigaci\'on en Computaci\'on, Instituto Polit\'ecnico Nacional, Mexico\\
$^{\ddagger}$School of Computer Science and Engineering, Nanyang Technological University, Singapore \\ 
$^{\Phi}$School of Computing, National University of Singapore, Singapore \\
$^{\bigtriangledown}$Computer Science \& Engineering, University of Michigan, Ann Arbor, USA\\
$^{\equiv}$ means authors contributed equally
%\\{\tt navo@nlp.cic.ipn.mx}, {\tt sporia@ntu.edu.sg}, {\tt hazarika@comp.nus.edu.sg}, \\ {\tt mihalcea@umich.edu}, {\tt gelbukh@gelbukh.com}, {\tt cambria@ntu.edu.sg}
}
\maketitle

\begin{abstract}

Emotion detection in conversations is a necessary step for a number of applications, including opinion mining
over chat history, social media threads, debates, argumentation mining, understanding consumer feedback
in live conversations, and so on. Currently systems do not treat the parties in the conversation individually by adapting to the speaker of each utterance. In this paper, we describe a new method based on recurrent neural networks that
keeps track of the individual party states throughout the conversation and uses this
information for emotion classification. Our model outperforms
the state-of-the-art by a significant margin on two different datasets. 
%Moreover, unlike
%the current state of the art, our model is scalable to multi-party scenarios without any modifications to
%the model and any increase in the number of parameters.

\end{abstract}

\section{Introduction}
\label{sec:introduction}
Emotion detection in conversations has been gaining increasing attention from the research community due to its applications in many
important tasks such as opinion mining over chat history and social media threads in YouTube,
Facebook, Twitter, and so on. %E-commerce giants like Amazon, BestBuy, Target has a lot to gain by making
%their stocking decisions based on user feedback and debates available online. Moreover, product designers
%can design or improve their products based on user emotion behind the different aspects of their products.
In this paper, we present a method based on recurrent neural networks (RNN) that can cater to these needs by
processing the huge amount of available conversational data.

\begin{figure}[h] 
	\centering 
 \small
	\includegraphics[width=\linewidth]{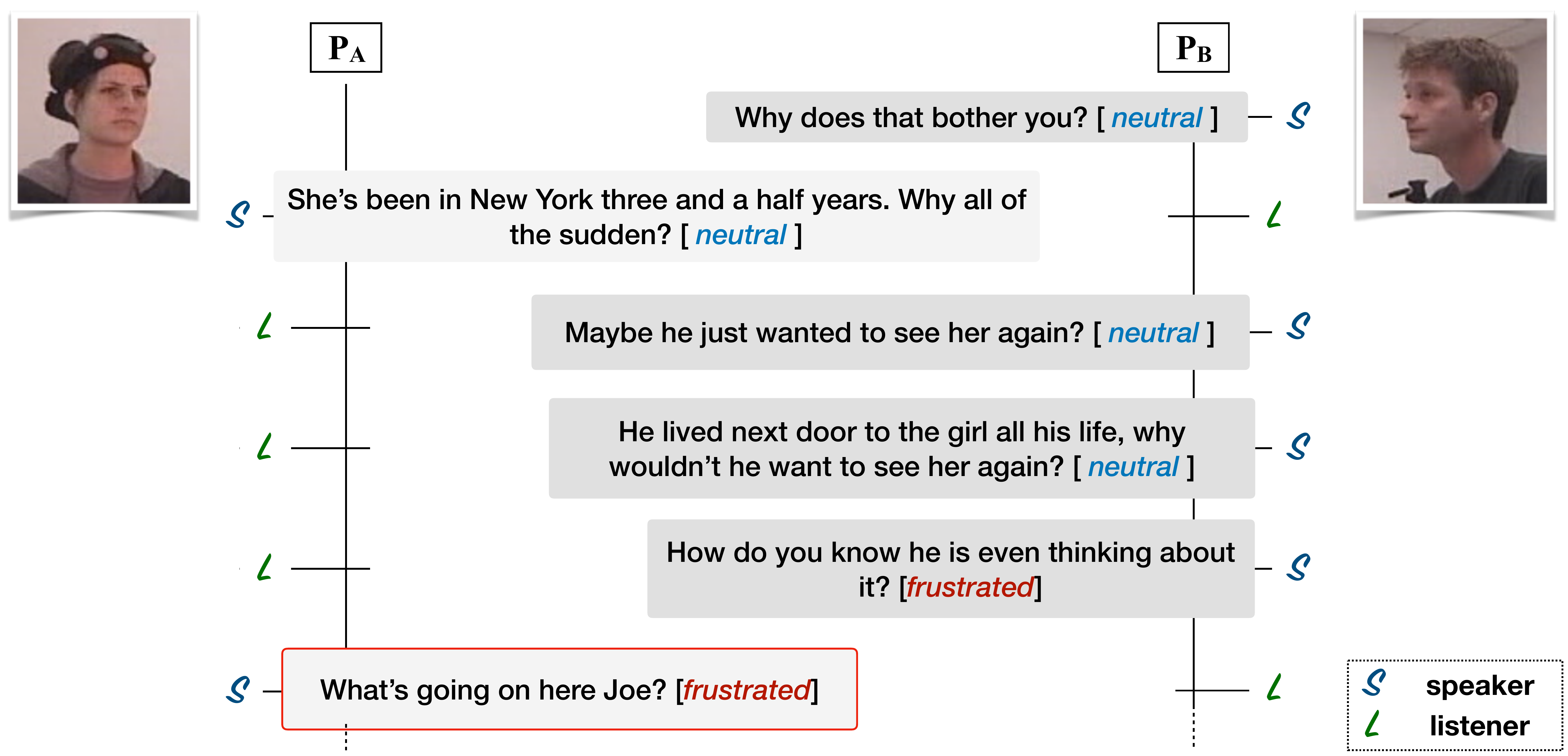} 
	\caption{ 
In this dialogue, $P_A$'s emotion changes are influenced by the behavior of $P_B$.}
	\label{fig:example}
\end{figure}

Current systems, including the state of the art~\cite{hazarika-EtAl:2018:N18-1}, do not distinguish different parties in a conversation in a meaningful way.
They are not aware of
the speaker of a given utterance.
In contrast, we model individual parties with party states, as the conversation flows,
by relying on the utterance, the context, and current party state.
Our model is based on the assumption that there are three major aspects relevant to the
emotion in a conversation: the speaker, the context from the preceding utterances, and the emotion of the preceding utterances.
These three aspects are not necessarily independent, but their separate modeling significantly
outperforms the state-of-the-art (\cref{tab:results-text}).
In dyadic conversations, the parties have distinct roles. Hence, to extract the context,
it is crucial to consider the preceding turns of both the speaker and the listener at a given
moment (\cref{fig:example}).
%DialogueRNN handles this by means of party state handling and connecting the states of different parties in a sequential manner with RNN.
%\hl{explain the assumption with an example}
%\hl{Speaker info provides better context}
%\hl{In conversation, inter-speaker dependency is key as shown in the figure}
%\hl{Not considering speaker info may result into noisy context and thus confuse the model}
%\\
%\hl{eta likhte hobe je in a dyadic conversation setting both the speakers play distinctive roles that are key to analyze .. at time t in the conversation, a speaker's emotion depend on the context which constitutes of that speaker's previous turns and the other user's previous turns ..}
%\\
%\hl{amader ekhane amra basically self history r other party's history
%dutor importance weight korchi
%ACL+att e seta hochhe kintu explicitly na .. just through lstm and attention 
%toh tor jodi erokom sequence thake PA, PB, PA, PB 
%sekhane PB just PA er theke info pachhe directly 
%but amra bolchi PA should directly get info from the previous state of PA
%etai better context e help korche 
%model at time t teh oi speaker er previous history ta remember korche r global state + other speaker er opor depend kore result dichhe }

Our proposed DialogueRNN system employs three gated recurrent units (GRU)~\cite{DBLP:journals/corr/ChungGCB14} to model
these aspects.
The incoming utterance is fed into two GRUs called global GRU and party GRU to update the context and party state, respectively. The global GRU encodes corresponding party information while encoding an utterance. 

Attending over this GRU gives contextual representation that has information of all preceding utterances by different parties in the conversation. The speaker state depends on this context through attention and the speaker's previous state. This
ensures that at time $t$, the speaker state directly gets information from the speaker's previous state and 
global GRU which has information on the preceding parties. Finally, the updated speaker state
is fed into the emotion GRU to decode the emotion representation of the given utterance, which is used
for emotion classification. At time $t$, the emotion GRU cell gets the emotion representation 
of $t-1$ and the speaker state of~$t$.

The emotion GRU, along with the global GRU, plays a 
pivotal role in inter-party relation modeling. On the other hand, party GRU models relation between two 
sequential states of the same party. In DialogueRNN, all these three different types of GRUs are connected in a 
recurrent manner. We believe that DialogueRNN outperforms state-of-the-art  contextual emotion classifiers such as \cite{hazarika-EtAl:2018:N18-1,poria-EtAl:2017:Long} because of better context representation.

The rest of the paper is organized as follows: \cref{sec:related-works} discusses related work;
\cref{sec:method} provides detailed description of our model; \cref{sec:experiments,sec:results-discussion} present the experimental results; finally, \cref{sec:conclusion}
concludes the paper.

% \hl{Hypothesis - the model should know the info if two utterances are uttered by different speakers.}
\section{Related Work}
\label{sec:related-works}

Emotion recognition has attracted attention in various fields such as natural language processing, psychology,
cognitive science, and so on \cite{5565330}. \citet{ekman1993facial} found correlation between emotion and
facial cues. 
\citet{datcu2008semantic} fused acoustic information with visual cues for emotion recognition.
\citet{alm2005emotions} introduced text-based emotion recognition, developed in the work of 
\citet{strapparava2010annotating}. \citet{wollmer2010context} used contextual information for emotion
recognition in multimodal setting. Recently, \citet{poria-EtAl:2017:Long}
successfully used RNN-based deep networks for multimodal emotion recognition, which was followed by
other works~\cite{chen2017multimodal,AAAI1817341,zadatt}. 

Reproducing human interaction requires deep understanding of conversation. \citet{ruusuvuori2013emotion} states that emotion plays a pivotal role in conversations.
It has been argued that emotional dynamics in a conversation is an inter-personal phenomenon~\cite{richards2003emotion}. Hence, our model incorporates inter-personal
interactions in an effective way. Further, since conversations have a natural temporal nature, we adopt
the temporal nature through recurrent network~\cite{poria-EtAl:2017:Long}.

Memory networks~\cite{Sukhbaatar:2015:EMN:2969442.2969512} has been successful in several NLP
areas, including question answering~\cite{Sukhbaatar:2015:EMN:2969442.2969512,kumar2016ask}, machine
translation~\cite{bahdanau2014neural}, speech recognition~\cite{graves2014neural}, and so on.
Thus, \citet{hazarika-EtAl:2018:N18-1} used memory networks for emotion recognition in dyadic conversations,
where two distinct memory networks enabled inter-speaker interaction, yielding state-of-the-art performance.

\section{Methodology}
\label{sec:method}
\subsection{Problem Definition}
\label{sec:problem-definition}

Let there be $M$ parties/participants $p_1,p_2,\dots,p_M$ ($M=2$ for the 
datasets we used) in a
conversation. The task is to predict the emotion labels (\textit{happy}, \textit{sad}, \textit{neutral},
\textit{angry}, \textit{excited}, and \textit{frustrated}) of the constituent utterances
$u_1,u_2,\dots,u_N$, where utterance $u_t$ is uttered by party $p_{s(u_t)}$,
while $s$ being the mapping between utterance and index of its corresponding
party. Also, $u_t\in \mathbb{R}^{D_m}$ is the utterance representation,
obtained using feature extractors %~(\cref{sec:feature-extraction}).
described below.

\subsection{Unimodal Feature Extraction}
\label{sec:feature-extraction}

% For the sake of a fair comparison with the state-of-the-art method CMN~\cite{hazarika-EtAl:2018:N18-1},
% we use identical unimodal features as \citet{hazarika-EtAl:2018:N18-1}.
For a fair comparison with the state-of-the-art method, conversational memory networks (CMN)~\cite{hazarika-EtAl:2018:N18-1},
we follow identical feature extraction procedures.

\subsubsection{Textual Feature Extraction}
\label{sec:text-feat-extr}

We employ convolutional neural networks (CNN) for textual feature extraction. Following
\citet{kim2014convolutional}, we obtain n-gram features from each utterance using three distinct
convolution filters
of sizes 3, 4, and 5 respectively, each having 50 feature-maps. Outputs are then subjected to max-pooling followed by rectified linear unit (ReLU) activation. These activations are concatenated and fed to a $100$ dimensional dense layer, which is regarded as the textual utterance representation. This network is trained at utterance level with the emotion labels.

\subsubsection{Audio and Visual Feature Extraction}
\label{sec:visu-feat-extr}
Identical to \citet{hazarika-EtAl:2018:N18-1}, we use 3D-CNN and openSMILE~\cite{eyben2010opensmile} for visual and acoustic feature extraction, respectively.

%\subsubsection{Acoustic Feature Extraction}
%\label{sec:acoust-feat-extr}

%Identical to \citet{hazarika-EtAl:2018:N18-1}, openSMILE~\cite{eyben2010opensmile} is used for
%acoustic feature extraction.

\subsection{Our Model}
\label{sec:model}

\begin{figure*}[ht]
  \centering
  %\small
  \begin{subfigure}{0.73\textwidth}
  \centering
  \includegraphics[width=\linewidth]{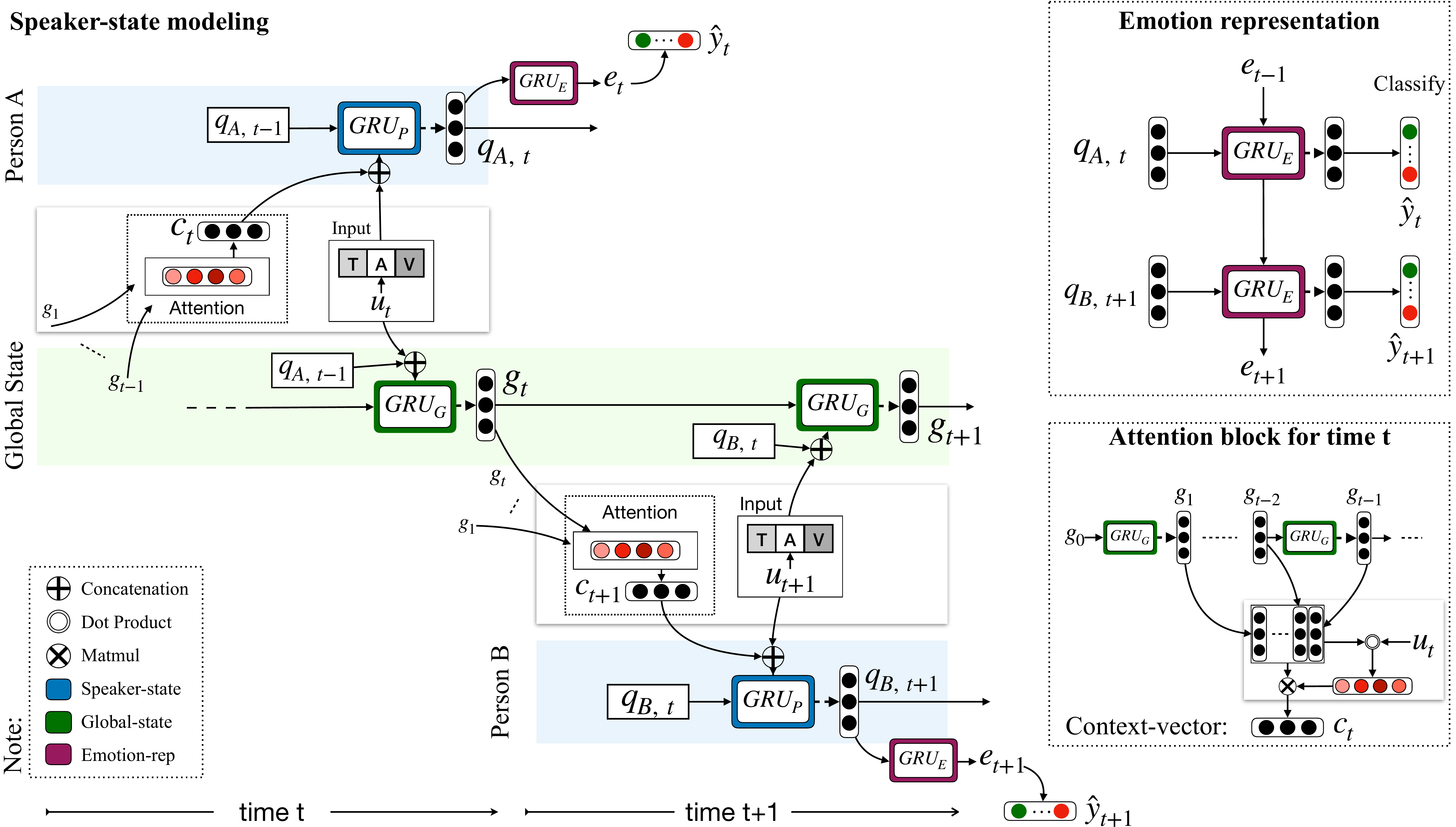}
  \caption{}
  \label{fig:architecture}
  \end{subfigure}
%   \hfill
  ~
	\begin{subfigure}{0.25\textwidth}
	\centering
  \small
	\includegraphics[width=\linewidth]{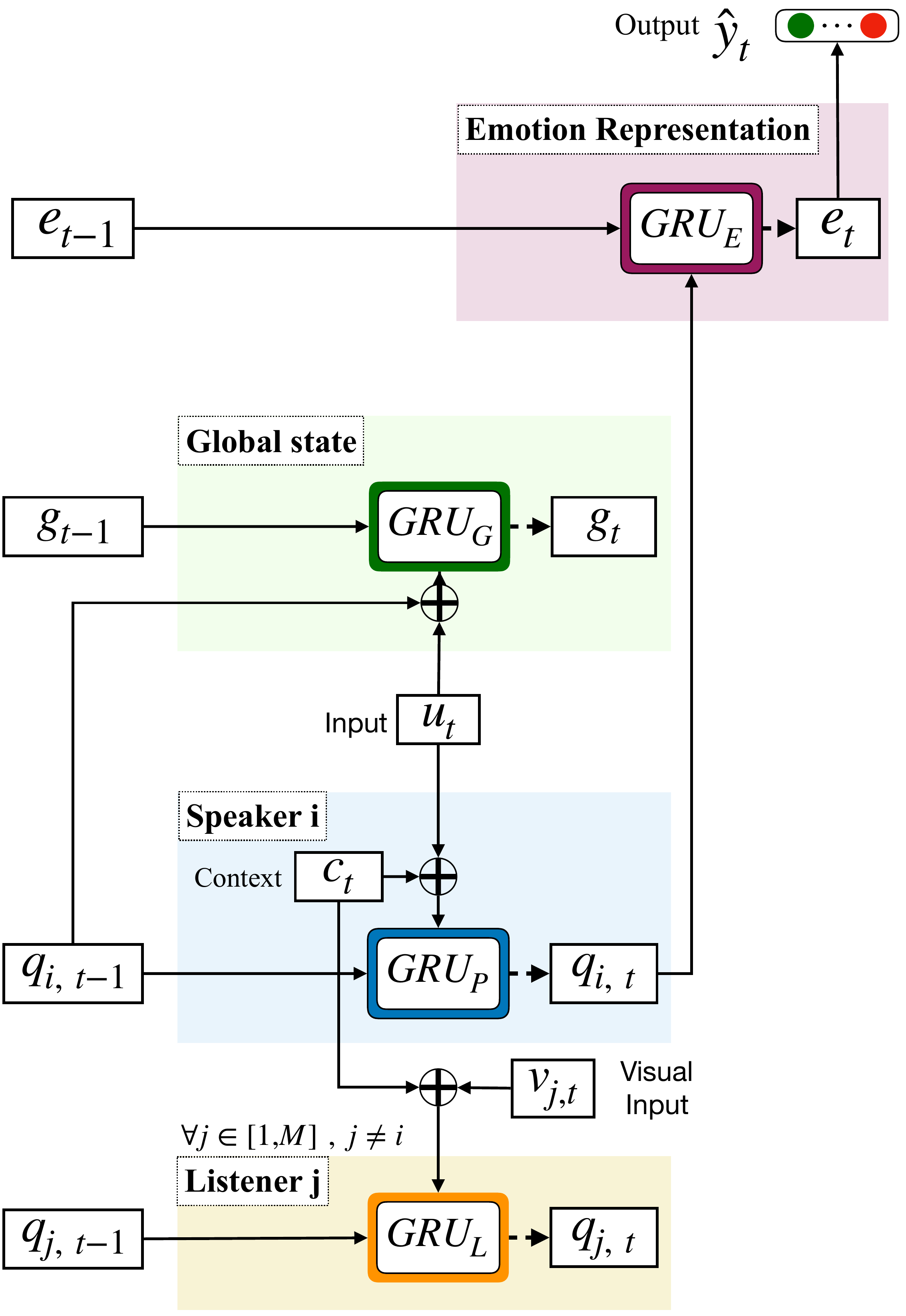}
	\caption{}
	\label{fig:StateUpdate}
	\end{subfigure}
\caption{(a) DialogueRNN architecture. (b) Update schemes for global, speaker, listener, and emotion states for $t^{th}$ utterance in a
	dialogue. Here, Person $i$ is the speaker and Persons $j \in [1, M] \ \text{and} \ j \neq i $ are the
	listeners.}
\end{figure*}

% teacher forcing
We assume that the emotion of an utterance in a conversation depends on three
major factors:
\begin{enumerate}
\item the speaker.
\item the context given by the preceding utterances.
\item the emotion behind the preceding utterances.
\end{enumerate}
Our model DialogueRNN,\footnote{Implementation available at
\url{https://github.com/senticnet/conv-emotion}}
shown in \cref{fig:architecture}, models these three factors as follows:
% using gated recurrent units (GRU)~\cite{DBLP:journals/corr/ChungGCB14}. 
each party is modeled using a \textit{party state} which changes as and when that party utters an utterance. This enables the model to track the parties' emotion dynamics 
through the conversations, which is related to the emotion behind the utterances.
% Furthermore, the context of an utterance is modeled using a \textit{global state}, where the preceding
% utterances and the speaker states are jointly encoded for context representation,
% necessary for accurate party state representation. Finally, the model infers emotion
% representation from the speaker party state along with the preceding speakers' state as
% context. This emotion representation is used for the final emotion classification.
Furthermore, the context of an utterance is modeled using a \textit{global state}
(called global, because of being shared among the parties), where the preceding
utterances and the party states are jointly encoded for context representation,
necessary for accurate party state representation. Finally, the model infers emotion
representation from the party state of the speaker along with the preceding speakers' states as
context. This emotion representation is used for the final emotion classification.

% We use GRU cells to update the aforementioned states and representations,
We use GRU cells~\cite{DBLP:journals/corr/ChungGCB14} to update the states and representations. Each GRU cell computes a hidden state defined as $h_t=GRU_{\ast}(h_{t-1},x_t)$, where
% \begin{flalign}
%   \label{eq:1}
%   r_{t}&=\sigma(W_{.,h}^rh_{t-1}+ W_{.,x}^rx_t+b_{.}^r),\\
%   z_{t}&=\sigma(W_{.,h}^zh_{t-1}+ W_{.,x}^zx_t+b_{.}^z),\\
%   c_t&=\tanh(W_{.,h}^c(r_t\circ h_{t-1})+ W_{.,x}^cx_t +b_{.}^c),\\
%   h_t&= z_t\circ h_{t-1}+(1-z_t)\circ c_t.
%   % i_{t}&=\sigma(W_{.}^i(h_{t-1}\oplus x_t)+b_{.}^i),\\
%   % o_{t}&=\sigma(W_{.}^o(h_{t-1}\oplus x_t)+b_{.}^o),\\
%   % f_{t}&=\sigma(W_{.}^f(h_{t-1}\oplus x_t)+b_{.}^f),\\
%   % \tilde{C_t}&=\tanh(W_{.}^c(h_{t-1}\oplus x_t)+b_{.}^c),\\
%   % C_t&=i_{t}\circ \tilde{C_t} + f_{t} \circ C_{t-1},\\
%   % h_t&=o_{t} \circ \tanh(C_t).
% \end{flalign}
$x_t$ is the current input and $h_{t-1}$ is the previous GRU state. $h_t$ also serves as the current GRU output. We provide the GRU computation details in the supplementary. GRUs are efficient networks with trainable parameters: $W_{\ast,\{h,x\}}^{\{r,z,c\}}$ and $b_{\ast}^{\{r,z,c\}}$. 

% \hl{GRU details are added to the supplementary material}.

We model the emotion representation of the current utterance as a function of
the emotion representation of the previous utterance and the state of the
current speaker. Finally, this emotion representation is sent to a softmax layer
for emotion classification.

% \begin{figure*}[ht]
%   \centering
%   \begin{subfigure}{0.7\textwidth}
%   \centering
%   \includegraphics[width=1.0\linewidth]{DialogueRNN.pdf}
%  \caption{}
%   \label{fig:architecture}
%   \end{subfigure}
%   ~
% \begin{subfigure}{0.20\textwidth}
% 	\centering
%   \small
% 	\includegraphics[width=1.0\linewidth]{StateUpdate}
% 	\caption{}
% 	\label{fig:StateUpdate}
% \end{subfigure}
% \caption{(a) DialogueRNN architecture. (b) Update schemes for global, speaker, listener and emotion states for $t^{th}$ utterance in a
% 	dialogue. Here, Person $i$ is the speaker and Persons $j \in [1, M] \ \text{and} \ j \neq i $ are the
% 	listeners.}
% \end{figure*}

\subsubsection{Global State (Global GRU)}
\label{sec:global-state}

Global state aims to capture the context of a given utterance by jointly encoding
utterance and speaker state. Each state also serves as speaker-specific utterance representation.
Attending on these states facilitates the inter-speaker and inter-utterance
dependencies to produce improved context representation. The current utterance $u_t$ changes the speaker's state from
$q_{s(u_t),t-1}$ to $q_{s(u_t),t}$. We capture this change with GRU cell
$GRU_{\mathcal{G}}$ with output size $D_{\mathcal{G}}$, using $u_t$ and
$q_{s(u_t),t-1}$:
\begin{flalign}
  \label{eq:3}
  g_t=GRU_{\mathcal{G}}(g_{t-1},(u_t\oplus q_{s(u_t),t-1})),
\end{flalign}
where $D_{\mathcal{G}}$ is the size of global state vector, $D_{\mathcal{P}}$ is
the size of party state vector, $W_{\mathcal{G},h}^{\{r,z,c\}}\in
\mathbb{R}^{D_{\mathcal{G}}\times D_{\mathcal{G}}}$, 
$W_{\mathcal{G},x}^{\{r,z,c\}}\in
\mathbb{R}^{D_{\mathcal{G}}\times (D_m+D_{\mathcal{P}})}$,
$b_{\mathcal{G}}^{\{r,z,c\}}\in \mathbb{R}^{D_{\mathcal{G}}}$,
$q_{s(u_t),t-1}\in \mathbb{R}^{D_{\mathcal{P}}}$, $g_t,g_{t-1}\in
\mathbb{R}^{D_{\mathcal{G}}}$, $D_{\mathcal{P}}$ is party state size,
and $\oplus$ represents concatenation.
% \hl{instead of vanilla GRU try gating on $u_t$ and $q_{s(u_t),t-1}$}

\subsubsection{Party State (Party GRU)}

DialogueRNN keeps track of the state of individual speakers using fixed size
vectors $q_1,q_2,\dots,q_M$ through out the conversation. These states are
representative of the speakers' state in the conversation, relevant to emotion
classification. We update these states based on the current (at time $t$) role
of a participant in the conversation, which is either speaker or
listener, and the incoming utterance $u_t$. These state vectors are initialized
with null vectors for all the participants. The main purpose of this module is to
ensure that the model is aware of the speaker of each utterance and handle it accordingly.

% \hl{don't know how to discern between paragraph and subsubsection; currently
%  with colon}

\subsubsection{Speaker Update (Speaker GRU):}

Speaker usually frames the response based on the context, which is the
preceding utterances in the conversation. Hence, we capture context $c_t$
relevant to the utterance $u_t$ as follows:
\begin{flalign}
  \alpha&=\softmax(u_t^TW_{\alpha}[g_1,g_2,\dots,g_{t-1}]),\label{eq:7}\\
  \softmax(x)&=[e^{x_1}/\Sigma_{i}e^{x_i},e^{x_2}/\Sigma_{i}e^{x_i},\dots],\\
  c_t&=\alpha[g_1,g_2,\dots,g_{t-1}]^T,\label{eq:6}
\end{flalign}
where $g_1,g_2,\dots,g_{t-1}$ are preceding $t-1$ global states
($g_i\in \mathbb{R}^{D_\mathcal{G}}$), $W_{\alpha}\in \mathbb{R}^{D_{m}\times
D_{\mathcal{G}}}$,
$\alpha^T\in \mathbb{R}^{(t-1)}$, and $c_t\in \mathbb{R}^{D_\mathcal{G}}$. In
\cref{eq:7}, we calculate attention scores $\alpha$ over the previous global
states representative of the previous utterances. This assigns higher attention
scores to the utterances emotionally relevant to $u_t$. Finally, in \cref{eq:6}
the context vector $c_t$ is calculated by pooling the previous global states
with $\alpha$.

Now, we employ a GRU cell $GRU_{\mathcal{P}}$ to update the current speaker state
$q_{s(u_t),t-1}$ to the new state $q_{s(u_t),t}$ based on incoming utterance
$u_t$ and context $c_t$ using GRU cell $GRU_{\mathcal{P}}$ of output size
$D_{\mathcal{P}}$
\begin{flalign}
  \label{eq:2}
  q_{s(u_t),t}&=GRU_{\mathcal{P}}(q_{s(u_t),t-1},(u_t\oplus c_t)),
\end{flalign}
where $W_{\mathcal{P},h}^{\{r,z,c\}}\in \mathbb{R}^{D_{\mathcal{P}}\times
D_{\mathcal{P}}}$, $W_{\mathcal{P},x}^{\{r,z,c\}}\in \mathbb{R}^{D_{\mathcal{P}}\times
(D_m+D_{\mathcal{G}})}$, $b_{\mathcal{P}}^{\{r,z,c\}}\in
\mathbb{R}^{D_{\mathcal{P}}}$, and $q_{s(u_t),t},q_{s(u_t),t-1}\in
\mathbb{R}^{D_{\mathcal{P}}}$.
This encodes the information on the current
utterance along with its context from the global GRU into the speaker's state
$q_{s(u_t)}$, which helps in emotion classification down the line.

% \hl{maybe consider the previous utterance as well; attention: add inter
% $\alpha_i$ linear combination, but problem is the size of $g$ is variable;
% instead of $W_{\alpha}$, use $u_t$}

\subsubsection{Listener Update:}

Listener state models the listeners' change of state due to the
speaker's utterance. We tried two listener state update mechanisms:
\begin{itemize}
\item Simply keep the state of the listener unchanged, that is
\begin{flalign}
  \forall i\neq s(u_t), q_{i,t}=q_{i,t-1}.
\end{flalign}
\item Employ another GRU cell $GRU_{\mathcal{L}}$ to update the listener state
  based on listener visual cues (facial expression) $v_{i,t}$ and its context
  $c_t$, as
  \begin{flalign}
    \label{eq:8}
    \forall i\neq s(u_t), q_{i,t}=GRU_{\mathcal{L}}(q_{i,t-1},(v_{i,t}\oplus c_t)),
  \end{flalign}
  where $v_{i,t}\in \mathbb{R}^{D_\mathcal{V}}$, $W_{\mathcal{L},h}^{\{r,z,c\}}\in \mathbb{R}^{D_{\mathcal{P}}\times
D_{\mathcal{P}}}$, $W_{\mathcal{L},x}^{\{r,z,c\}}\in \mathbb{R}^{D_{\mathcal{P}}\times
(D_{\mathcal{V}}+D_{\mathcal{G}})}$, and $b_{\mathcal{L}}^{\{r,z,c\}}\in \mathbb{R}^{D_{\mathcal{P}}}$.
Listener visual features of party $i$ at time $t$ $v_{i,t}$ are extracted using the model introduced by~\citet{DBLP:journals/corr/abs-1710-07557}, pretrained on FER2013 dataset, where feature size $D_{\mathcal{V}}=7$.
\end{itemize}
The simpler first approach turns out to be sufficient, since the second approach
yields very similar result while increasing number of parameters. This is due to
the fact that a listener becomes relevant to the conversation only when he/she
speaks. In other words, a silent party has no influence in a conversation. Now, when a
party speaks, we update his/her state $q_i$ with context $c_t$ which contains
relevant information on all the preceding utterances, rendering explicit
listener state update unnecessary. This is shown in
\cref{tab:results-text}.

\subsubsection{Emotion Representation (Emotion GRU)}

We infer the emotionally relevant representation $e_t$ of utterance $u_t$ from
the speaker's state $q_{s(u_t),t}$ and the emotion representation of the
previous utterance $e_{t-1}$. Since context is important to the emotion of the
incoming utterance $u_t$, $e_{t-1}$ feeds fine-tuned emotionally relevant
contextual information from other the party states $q_{s(u_{<t}),<t}$ into the
emotion representation $e_t$. This establishes a connection between the speaker 
state and the other party states. Hence, we model $e_t$ with a GRU cell
($GRU_{\mathcal{E}}$) with output size $D_{\mathcal{E}}$ as
\begin{flalign}
  \label{eq:4}
  e_t=GRU_{\mathcal{E}}(e_{t-1},q_{s(u_t),t}),
\end{flalign}
where $D_{\mathcal{E}}$ is the size of emotion representation vector,
$e_{\{t,t-1\}}\in \mathbb{R}^{D_{\mathcal{E}}}$, $W_{\mathcal{E},h}^{\{r,z,c\}}\in
\mathbb{R}^{D_{\mathcal{E}}\times D_{\mathcal{E}}}$, $W_{\mathcal{E},x}^{\{r,z,c\}}\in
\mathbb{R}^{D_{\mathcal{E}}\times D_{\mathcal{P}}}$, and
$b_{\mathcal{E}}^{\{r,z,c\}}\in \mathbb{R}^{D_{\mathcal{E}}}$.

Since speaker state gets information from global states, which serve as speaker-specific utterance
representation, one may claim that this way the model already has access to the information on other
parties. However, as shown in the ablation study (\cref{sec:ablation-study}) emotion GRU helps to improve the
performance by directly linking states of preceding parties. Further, we believe that speaker and global GRUs
($GRU_{\mathcal{P}}$, $GRU_{\mathcal{G}}$) jointly act similar to an encoder, whereas emotion GRU serves as a
decoder.

\subsubsection{Emotion Classification}

We use a two-layer perceptron with a final softmax layer to calculate $c=6$
emotion-class probabilities from emotion representation $e_t$ of utterance $u_t$
and then we pick the most likely emotion class:
\begin{flalign}
  l_t&=\ReLU(W_{l}e_t+b_{l}),\label{eq:5}\\
  \mathcal{P}_t&=\softmax(W_{smax}l_t+b_{smax}),\label{eq:c-6}\\
  \hat{y_t}&=\argmax_{i}(\mathcal{P}_t[i]),\label{eq:c-7}
\end{flalign}
where $W_l\in \mathbb{R}^{D_l\times D_\mathcal{E}}$, $b_l\in \mathbb{R}^{D_l}$,
$W_{smax}\in \mathbb{R}^{c\times D_{l}}$, $b_{smax}\in
\mathbb{R}^{c}$, $\mathcal{P}_t\in \mathbb{R}^c$, and $\hat{y_t}$ is the
predicted label for utterance $u_t$.

\subsubsection{Training}
\label{sec:training}

We use categorical cross-entropy along with L2-regularization as the measure of
loss ($L$) during training:
\begin{flalign}
  \label{eq:9}
  L=-\frac{1}{\sum_{s=1}^Nc(s)}\sum_{i=1}^N \sum_{j=1}^{c(i)}\log
  \mathcal{P}_{i,j}[y_{i,j}]+\lambda \norm{\theta}_2,
\end{flalign}
where $N$ is the number of samples/dialogues, $c(i)$ is the number of utterances
in sample $i$, $\mathcal{P}_{i,j}$ is the probability distribution of emotion
labels for utterance $j$ of dialogue $i$, $y_{i,j}$ is the expected class label
of utterance $j$ of dialogue $i$, $\lambda$ is the L2-regularizer weight, and
$\theta$ is the set of trainable parameters where
\begin{flalign*}
  \theta=\{&W_{\alpha},W_{\mathcal{P},\{h,x\}}^{\{r,z,c\}},b_{\mathcal{P}}^{\{r,z,c\}},W_{\mathcal{G},\{h,x\}}^{\{r,z,c\}},b_{\mathcal{G}}^{\{r,z,c\}},W_{\mathcal{E},\{h,x\}}^{\{r,z,c\}},\\
  &b_{\mathcal{E}}^{\{r,z,c\}},W_l,b_l,W_{smax},b_{smax}\}.
\end{flalign*}

We used stochastic gradient descent based Adam~\cite{DBLP:journals/corr/KingmaB14} optimizer to
train our network. Hyperparameters are optimized using grid search (values are added to the supplementary material).

% \begin{flalign*}
%   g_t&=GRU_{\mathcal{G}}(g_{t-1},(u_t\oplus q_{s(u_t),t-1})),\\
%   c_t&=Attn([g_1,g_2,\dots,g_{t-1}]),\\
%   q_{s(u_t),t}&=GRU_{\mathcal{P}}(q_{s(u_t),t-1},(u_t\oplus c_t)),\\
%   q_{i,t}&=q_{i,t-1}, \forall i\neq s(u_t),\\
%   e_t&=GRU_{\mathcal{E}}(e_{t-1},q_{s(u_t),t}),\\
%   \hat{y_t}&=\argmax_{i}(\softmax(FC(e_t))_{i=1}^{c}).
% \end{flalign*}

\subsection{DialogueRNN Variants}
\label{sec:dialoguernn-variants}

We use DialogueRNN~(\cref{sec:model}) as the basis for the following models:

\subsubsection{DialogueRNN + Listener State Update (DialogueRNN$_l$):}

This variant updates the listener state based on the the resulting speaker state
$q_{s(u_t),t}$, as described in \cref{eq:8}.
\subsubsection{Bidirectional DialogueRNN (BiDialogueRNN):}

Bidirectional DialogueRNN is analogous to bidirectional RNNs, where two different RNNs are used for forward and backward passes of the input sequence. Outputs from the RNNs are concatenated in sequence level. Similarly, in BiDialogueRNN, the final emotion representation contains information from both past and future utterances in the dialogue through forward and backward DialogueRNNs respectively, which provides better context for emotion classification.

\subsubsection{DialogueRNN + attention (DialogueRNN+Att):}

For each emotion representation $e_t$, attention is applied over all
surrounding emotion representations in the dialogue by matching them with $e_t$ (\cref{eq:beta,eq:beta-2}). This
provides context from the relevant (based on attention score) future and
preceding utterances.

\subsubsection{Bidirectional DialogueRNN + Emotional attention (BiDialogueRNN+Att):}

For each emotion representation $e_t$ of BiDialogueRNN, attention is applied over all the emotion representations
in the dialogue to capture context from the other utterances in dialogue:
\begin{flalign}
  \beta_t&=\softmax(e_t^T W_{\beta} [e_1, e_2,\dots,e_N]), \label{eq:beta}\\
  \tilde{e}_t&=\beta_t[e_1, e_2, \dots, e_N]^T, \label{eq:beta-2}
\end{flalign}
where $e_t\in \mathbb{R}^{2D_\mathcal{E}}$, $W_\beta \in \mathbb{R}^{2D_\mathcal{E}\times 2D_\mathcal{E}}$,
$\tilde{e}_t\in \mathbb{R}^{2D_\mathcal{E}}$, and $\beta_t^T\in \mathbb{R}^N$. Further, $\tilde{e}_t$ are fed
to a two-layer perceptron for emotion classification, as in \cref{eq:5,eq:c-6,eq:c-7}.

\section{Experimental Setting}
\label{sec:experiments}

\subsection{Datasets Used}
\label{sec:dataset-details}

We use two emotion detection datasets IEMOCAP~\cite{iemocap} and
AVEC~\cite{Schuller:2012:ACA:2388676.2388776} to evaluate DialogueRNN. We partition
both datasets into train and test sets with roughly $80/20$ ratio
such that the partitions do not share any speaker. \cref{table:dataset} shows
the distribution of train and test samples for both dataset.

\begin{table}[h]
\small
\begin{center}
\begin{tabular}{|c|c|c|c|}
\hline
\multirow{2}{*}{Dataset}&\multirow{2}{*}{Partition}& Utterance & Dialogue\\
&& Count & Count\\
\hline
\hline
\multirow{2}{*}{IEMOCAP}&train + val&5810&120\\
\cline{2-4}&test&1623&31\\
\hline
\multirow{2}{*}{AVEC}&train + val&4368&63\\
\cline{2-4}&test&1430&32\\
\hline
\end{tabular}
\end{center}
\caption{Dataset split ((train + val) / test $\approx 80\%/20\%$).}
\label{table:dataset}
\end{table}

\subsubsection{IEMOCAP:}

IEMOCAP~\cite{iemocap} dataset contains videos of two-way conversations of ten
unique speakers, where only the first eight speakers from session one to four
belong to the train-set. Each video contains a single dyadic dialogue, segmented
into utterances. The utterances are annotated with one of six emotion labels,
which are happy, sad, neutral, angry, excited, and frustrated.

\subsubsection{AVEC:}

AVEC~\cite{Schuller:2012:ACA:2388676.2388776} dataset is a modification of
SEMAINE database~\cite{5959155} containing interactions between humans and
artificially intelligent agents. Each utterance of a dialogue is annotated with
four real valued affective attributes: valence ($[-1,1]$), arousal ($[-1,1]$),
expectancy ($[-1,1]$), and power ($[0,\infty)$). The annotations are available
every 0.2 seconds in the original database. However, in order to adapt the
annotations to our need of utterance-level annotation, we averaged the
attributes over the span of an utterance.

\begin{table*}[ht!]
%\begin{tabular}{llc}
%Hello & Soujanya  & 1234\\
%This is & the size& 5678\\
%Hello & Soujanya  & 1234\\
%This is & the size& 5678\\
%Hello & Soujanya  & 1234\\
%This is & the size& 5678
%\end{tabular}
%\small
  \centering
  \resizebox{\linewidth}{!}{
   \begin{tabular}{|l||c@{~~}c|c@{~~}c|c@{~~}c|c@{~~}c|c@{~~}c|c@{~~}c|c@{~~}c||c@{~~}c|c@{~~}c|c@{~~}c|c@{~~}c|}
    \hline
    \multirow{3}{*}{Methods} & \multicolumn{14}{c||}{IEMOCAP} & \multicolumn{8}{c|}{AVEC}\\
    \cline{2-23} & \multicolumn{2}{c|}{Happy} & \multicolumn{2}{c|}{Sad} &
                                        \multicolumn{2}{c|}{Neutral} & \multicolumn{2}{c|}{Angry} & \multicolumn{2}{c|}{Excited} & \multicolumn{2}{c|}{Frustrated} & \multicolumn{2}{c||}{\textbf{Average(w)}}&\multicolumn{2}{c|}{Valence}& \multicolumn{2}{c|}{Arousal}& \multicolumn{2}{c|}{Expectancy} & \multicolumn{2}{c|}{Power}\\
    \cline{2-23} & Acc. & F1 & Acc. & F1 & Acc. & F1 & Acc. & F1 & Acc. & F1 & Acc. & F1 & Acc. & F1 & $MAE$ & $r$ & $MAE$ & $r$ & $MAE$ & $r$ & $MAE$ & $r$\\
    \hline
    \hline
  CNN &27.77&29.86&57.14&53.83&34.33&40.14&61.17&52.44&46.15&50.09&62.99&55.75&48.92&48.18&0.545&-0.01&0.542&0.01&0.605&-0.01&8.71&0.19 \\
    memnet &25.72&33.53&55.53&61.77&58.12&52.84&59.32&55.39&51.50&58.30&67.20&59.00&55.72&55.10&0.202&0.16&0.211&0.24&0.216&0.23&8.97&0.05\\
    c-LSTM &29.17&34.43&57.14&60.87&54.17&51.81&57.06&56.73&51.17&57.95&67.19&58.92&55.21&54.95&0.194&0.14&0.212&0.23&0.201&0.25&8.90&-0.04 \\
    c-LSTM+Att &30.56&35.63&56.73&62.90&57.55&53.00&59.41&59.24&52.84&58.85&65.88&59.41&56.32&56.19&0.189&0.16&0.213&0.25&0.190&0.24&8.67&0.10 \\
    \hline
        CMN (SOTA) &25.00&30.38&55.92&62.41&52.86&52.39&61.76&59.83&55.52&60.25&71.13&60.69&56.56&56.13&0.192&0.23&0.213&0.29&0.195&0.26&8.74&-0.02 \\
    \hline
    DialogueRNN &31.25&33.83&66.12&69.83&63.02& 57.76&61.76&62.50&61.54&64.45&59.58&59.46&59.33&59.89&0.188&0.28&0.201&0.36&0.188&0.32&8.19&0.31 \\
    DialogueRNN$_l$ &35.42&35.54&65.71&69.85&55.73& 55.30&62.94&61.85&59.20&62.21&63.52&59.38&58.66&58.76&0.189&0.27&0.203&0.33&0.188&0.30&8.21&0.30 \\
    BiDialogueRNN &32.64&36.15&71.02&74.04&60.47&56.16&62.94&63.88&56.52&62.02&65.62&{\bf 61.73}&60.32&60.28&0.181&0.30&0.198&0.34&0.187&0.34&8.14&0.32\\
    DialogueRNN+Att &28.47&{\bf 36.61}&65.31&72.40&62.50&57.21&67.65&{\bf 65.71}&70.90&68.61&61.68&60.80&61.80&61.51&0.173&0.35&0.168&0.55&0.177&0.37&7.91&0.35\\
    BiDialogueRNN+Att &25.69&33.18&75.10&{\bf 78.80}&58.59&{\bf 59.21}&64.71&65.28&80.27&{\bf 71.86}&61.15&58.91&63.40&{\bf 62.75}&{\bf 0.168}&{\bf 0.35}&{\bf 0.165}&{\bf 0.59}&{\bf 0.175}&{\bf 0.37}&{\bf 7.90}&{\bf 0.37}\\
    \hline
   \end{tabular}
  }
  \caption{Comparison with the baseline methods for textual modality; Acc. = Accuracy,
  $MAE$ = Mean Absolute Error, $r$ = Pearson correlation coefficient; bold font denotes
  the best performances. Average(w) = Weighted average.}
  \label{tab:results-text}
\end{table*}
\begin{table}[ht]
  \centering
  \small
  \resizebox{\linewidth}{!}%
{%
\tabcolsep=3pt
\newcommand\stack[2]{$\stackrel{\mathrm{\displaystyle #1}}{#2}$}
\begin{tabular}{|c|c|c|c|c|c|}
    \hline
     \multirow{2}{*}{Methods} & IEMOCAP & \multicolumn{4}{c|}{AVEC} \\
     \cline{2-6} & F1 
     & \stack{Valence}{(r)}
     & \stack{Arousal}{(r)}
     & \stack{Expectancy}{(r)}
     & \stack{Power}{(r)}\\
    \hline
     TFN & 56.8 & 0.01 & 0.10 & 0.12 & 0.12\\
     MFN & 53.5 & 0.14& 25 & 0.26 & 0.15\\
     c-LSTM & 58.3 & 0.14 & 0.23 & 0.25 & -0.04\\
     CMN & 58.5 & 0.23 & 0.30 & 0.26 & -0.02\\
     \hline
     BiDialogueRNN+att$_{text}$ & 62.7 & 0.35 & 0.59 & 0.37 &
     0.37\\
     BiDialogueRNN+att$_{MM}$ & {\bf 62.9} & {\bf 0.37} & {\bf 0.60} & {\bf 0.37} & {\bf 0.41}\\
     \hline
  \end{tabular}
  }
  \caption{Comparison with the baselines for trimodal (T+V+A) scenario. BiDialogueRNN+att$_{MM}$ = BiDialogueRNN+att in multimodal setting.}
  %\vspace{-0.4cm}
  \label{tab:results-multimodal}
\end{table}

\subsection{Baselines and State of the Art}
\label{sec:baselines}

For a comprehensive evaluation of DialogueRNN, we compare our model with the following baseline methods:

\subsubsection{c-LSTM~\cite{poria-EtAl:2017:Long}:}

Biredectional LSTM \cite{hochreiter1997long} is used to capture the context from
the surrounding utterances to generate context-aware utterance
representation. However, this model does not differentiate among the speakers.

\subsubsection{c-LSTM+Att~\cite{poria-EtAl:2017:Long}:}

In this variant attention is applied applied to the c-LSTM output at each timestamp by following
\cref{eq:beta,eq:beta-2}. This provides better context to the final utterance representation.

\subsubsection{TFN~\cite{zadeh-EtAl:2017:EMNLP2017}:}

This is specific to multimodal scenario.
Tensor outer product is used to capture inter-modality and intra-modality
interactions. This model does not capture context from surrounding utterances.

\subsubsection{MFN~\cite{AAAI1817341}:}

Specific to multimodal scenario, this model utilizes multi-view learning by modeling view-specific and cross-view interactions. Similar to TFN, this model does not use contextual information.

\subsubsection{CNN~\cite{kim2014convolutional}:}
%This is only for textual modality. 
This is identical to our textual feature extractor network (\cref{sec:feature-extraction}) and it does not use contextual information from the surrounding utterances. %Hence, lower performance is expected.

\subsubsection{Memnet~\cite{Sukhbaatar:2015:EMN:2969442.2969512}:}

As described in \citet{hazarika-EtAl:2018:N18-1}, the current utterance is fed to a
memory network, where the memories correspond to preceding utterances.
The output from the memory network is used as the final utterance representation for
emotion classification.

\subsubsection{CMN~\cite{hazarika-EtAl:2018:N18-1}:}

This state-of-the-art method models utterance context from dialogue history
using two distinct GRUs for two speakers. Finally, utterance representation
is obtained by feeding the current utterance as query to two distinct memory
networks for both speakers.

\subsection{Modalities}
\label{sec:exper-sett}

We evaluated our model primarily on textual modality. However, to substantiate
efficacy of our model in multimodal scenario, we also experimented with multimodal features.

\section{Results and Discussion}
\label{sec:results-discussion}

We compare DialogueRNN and its variants with the baselines for textual data in
\cref{tab:results-text}. As expected, on average DialogueRNN outperforms all the
baseline methods, including the state-of-the-art CMN, on both of the
datasets.

\subsection{Comparison with the State of the Art}

We compare the performance of DialogueRNN against the performance of the state-of-the-art 
CMN on IEMOCAP and AVEC datasets for textual modality.

\subsubsection{IEMOCAP}

As evidenced by \cref{tab:results-text}, for IEMOCAP dataset, our model surpasses
the state-of-the-art method CMN by $2.77\%$ accuracy and $3.76\%$ f1-score on average.
We think that this enhancement is caused by the fundamental differences between CMN and
DialogueRNN, which are
\begin{enumerate}
  \item party state modeling with $GRU_{\mathcal{P}}$ in \cref{eq:2},
  \item speaker specific utterance treatment in \cref{eq:2,eq:3},
  \item and global state capturing with $GRU_{\mathcal{G}}$ in \cref{eq:3}.
\end{enumerate}

Since we deal with six unbalanced emotion labels, we also explored the
model performance for individual labels. DialogueRNN outperforms the state-of-the-art method
CMN in five out of six emotion classes by significant margin. For \emph{frustrated} class,
DialogueRNN lags behind CMN by $1.23\%$ f1-score. We think that DialogueRNN may surpass CMN
using a standalone classifier for \emph{frustrated} class. However, it can be observed in 
\cref{tab:results-text} that some of the other variants of DialogueRNN, like BiDialogueRNN has already outperformed CMN for \emph{frustrated} class.

\subsubsection{AVEC}

DialogueRNN outperforms CMN for valence, arousal, expectancy, and
power attributes; see \cref{tab:results-text}. It yields significantly lower mean absolute error ($MAE$)
and higher Pearson correlation coefficient ($r$) for all four attributes. We believe this to be due to
the incorporation of party state and emotion GRU, which are missing from CMN.

\subsection{DialogueRNN vs. DialogueRNN Variants}

%We discuss the performance improvement and deterioration of DialogueRNN variants on IEMOCAP and
%AVEC datasets for textual modality.
We discuss the performance of different DialogueRNN variants on IEMOCAP and AVEC datasets for textual modality.

\subsubsection{DialogueRNN$_l$:}

Following \cref{tab:results-text}, using explicit listener state update yields slightly worse
performance than regular DialogueRNN. This is true for both IEMOCAP and AVEC datasets in general.
However, the only exception to this trend is for \emph{happy} emotion label for IEMOCAP, where
DialogueRNN$_l$ outperforms DialogueRNN by $1.71\%$ f1-score. We surmise that, this is due to
the fact that a listener becomes relevant to the conversation only when he/she
speaks. Now, in DialogueRNN, when a party speaks, we update his/her state $q_i$ with context
$c_t$ which contains relevant information on all the preceding utterances, rendering
explicit listener state update of DialogueRNN$_l$ unnecessary.

\subsubsection{BiDialogueRNN:}

Since BiDialogueRNN captures context from the future utterances, we expect improved
performance from it over DialogueRNN. This is confirmed in \cref{tab:results-text}, where
BiDialogueRNN outperforms DialogueRNN on average on both datasets.

\subsubsection{DialogueRNN+Attn:}

%Similar to BiDialogueRNN, 
DialogueRNN+Attn also uses information from the future utterances.
However, here we take information from both past and future utterances by matching them with
the current utterance and calculating attention score over them. This provides relevance to 
emotionally important context utterances, yielding better performance than BiDialogueRNN.
The improvement over BiDialogueRNN is $1.23\%$ f1-score for IEMOCAP and consistently
lower $MAE$ and higher $r$ in AVEC.

\subsubsection{BiDialogueRNN+Attn:}

Since this setting generates the final emotion representation by attending over the
emotion representation from BiDialogueRNN, we expect better performance than both BiDialogueRNN and
DialogueRNN+Attn. This is confirmed in \cref{tab:results-text}, where this setting performs the best
in general than any other methods discussed, on both datasets. This setting yields $6.62\%$ higher
f1-score on average than the state-of-the-art CMN and $2.86\%$ higher f1-score than vanilla DialogueRNN 
for IEMOCAP dataset. For AVEC dataset also, this setting gives the best performance across all the four
attributes.

\subsection{Multimodal Setting}
As both IEMOCAP and AVEC dataset contain multimodal information, we have evaluated DialogueRNN on multimodal
features as used and provided by \citet{hazarika-EtAl:2018:N18-1}. We use concatenation of the unimodal features as a fusion method by following \citet{hazarika-EtAl:2018:N18-1}, since fusion mechanism is not a focus of this paper.
% We simply concatenate the unimodal features
% to obtain multimodal features, since fusion mechanism is not a focus of this paper.
Now, as we can see in \cref{tab:results-multimodal}, DialogueRNN significantly outperforms the strong baselines and state-of-the-art method CMN.

\subsection{Case Studies}
\label{sec:case-study}

\subsubsection{Dependency on preceding utterances (DialogueRNN)}

One of the crucial components of DialogueRNN is its attention module over the outputs of global GRU
($GRU_{\mathcal{G}}$).  Figure~\ref{fig:AlphaAttention}  shows the $\alpha$ attention vector~(\cref{eq:7}) over
the history of a given test utterance  compared with the attention vector from the CMN model. The attention of
our model is more focused compared with CMN: the latter gives diluted attention scores leading to misclassifications.
We observe this trend of focused-attention across cases and posit that it can be interpreted as a confidence indicator. Further in this example, the test utterance by $P_A$ (turn 44) comprises of a change in emotion from \textit{neutral} to \textit{frustrated}. DialogueRNN anticipates this correctly by attending to turn 41 and 42 that are spoken by $P_A$ and $P_B$, respectively. These two utterances provide self and inter-party influences that trigger the emotional shift. CMN, however, fails to capture such dependencies and wrongly predicts \textit{neutral} emotion.
%It is also apparent that in DialogueRNN, during emotion shifts, the co-party's turns get stronger attention over its own previous turn.

%\begin{figure}[t] 
%	\centering 
%  \small
%	\includegraphics[width=0.8\linewidth]{alpha_attention} 
%	\caption[]{Comparison of attention scores over utterance history of CMN and DialogueRNN}
%	\label{fig:AlphaAttention}
%\end{figure}

\subsubsection{Dependency on future utterances (BiDialogueRNN+Att)}
%\begin{figure}[b]
%  \centering
%  \includegraphics[width=0.9\linewidth]{future_attention.pdf}
%  \caption{Illustration of the $\beta$ attention weights over emotion representations $e_t$ for a
%  segment of conversation between a couple; $P_A$ = the woman, $P_B$ = the man.}
%  \label{fig:case-study}
%\end{figure}

\begin{figure*}[!htbp]
  \centering
  \begin{subfigure}{0.49\textwidth}
  \centering
	\includegraphics[width=\linewidth]{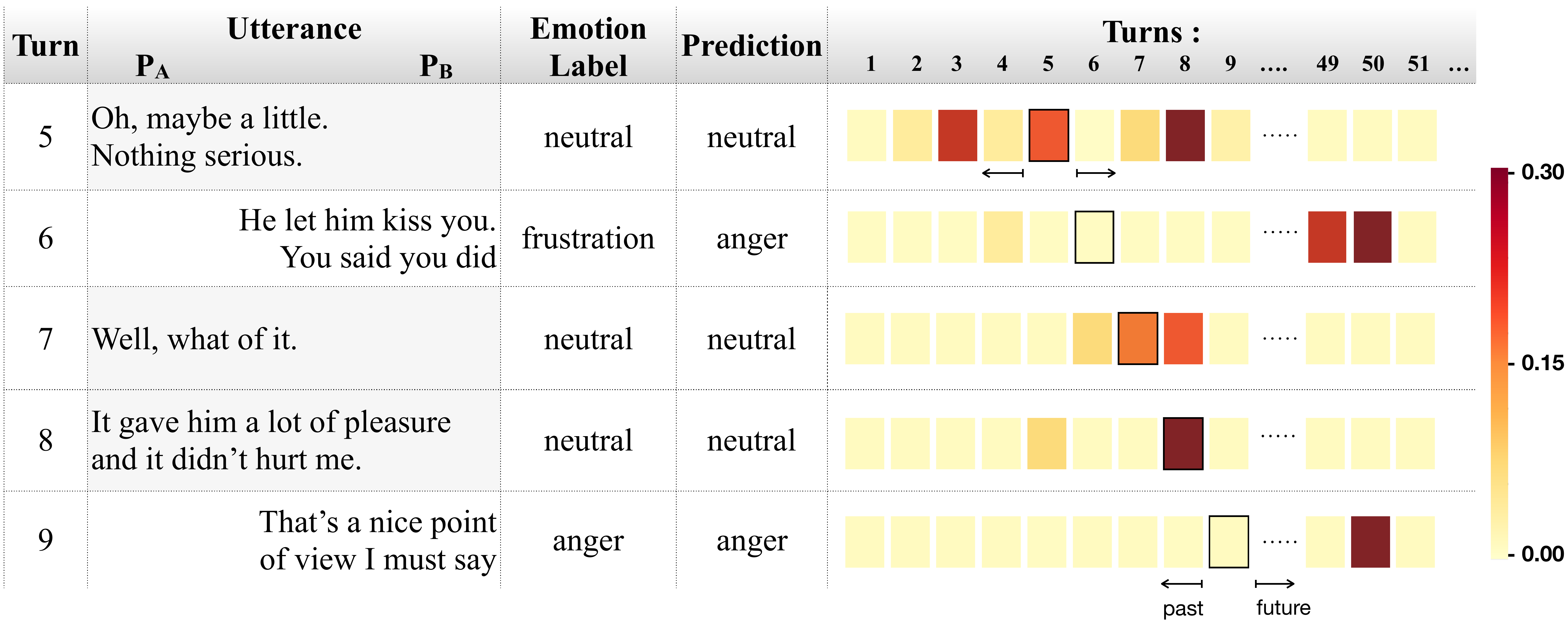}
  \caption{}
  \label{fig:case-study}
  \end{subfigure}%\hspace{0.8cm}
  ~
\begin{subfigure}{0.49\textwidth}
	\centering
  \small
	\includegraphics[width=\linewidth]{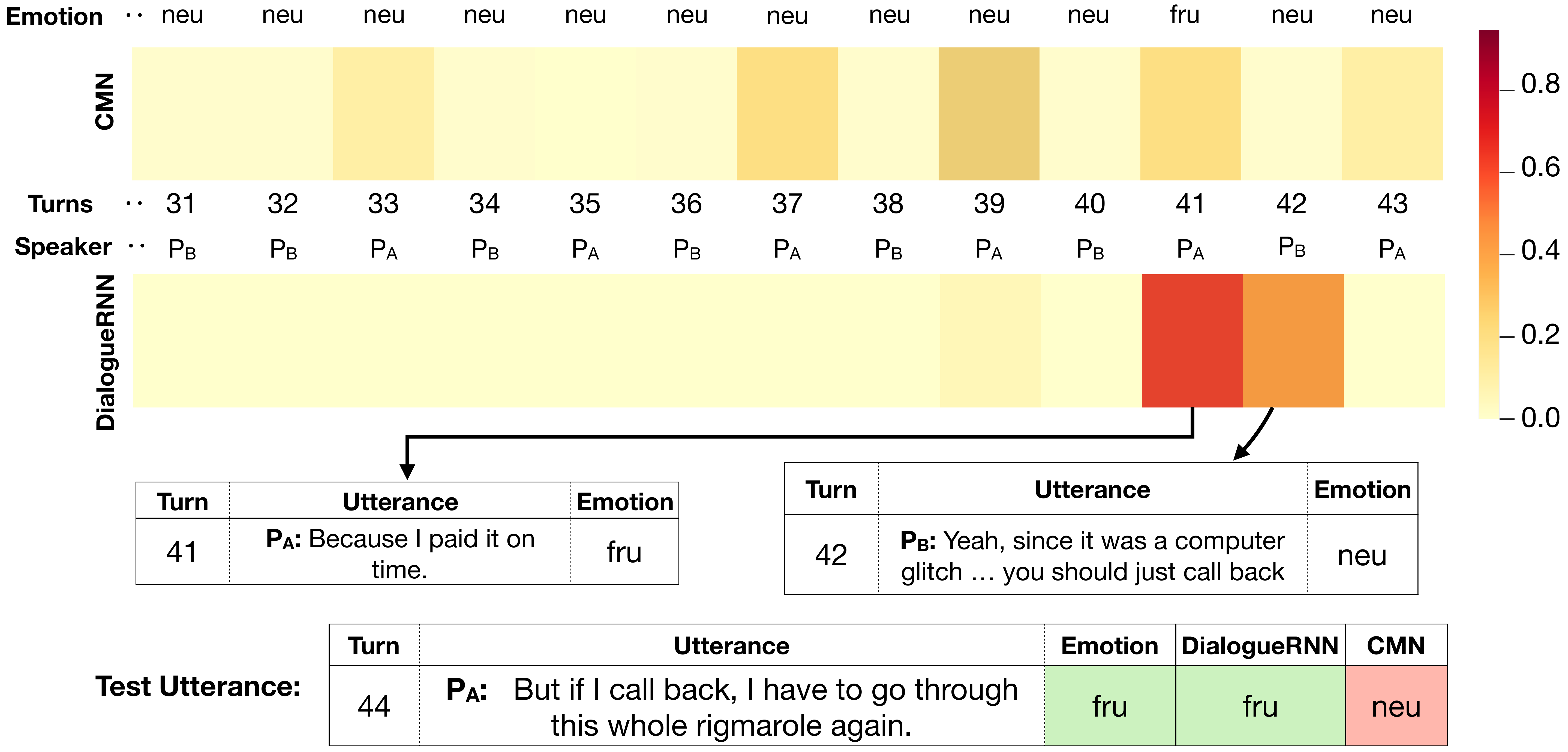} 
	\caption{}
	\label{fig:AlphaAttention}
\end{subfigure}
\begin{subfigure}{0.49\textwidth}
	\centering
  \small
	\includegraphics[width=\linewidth]{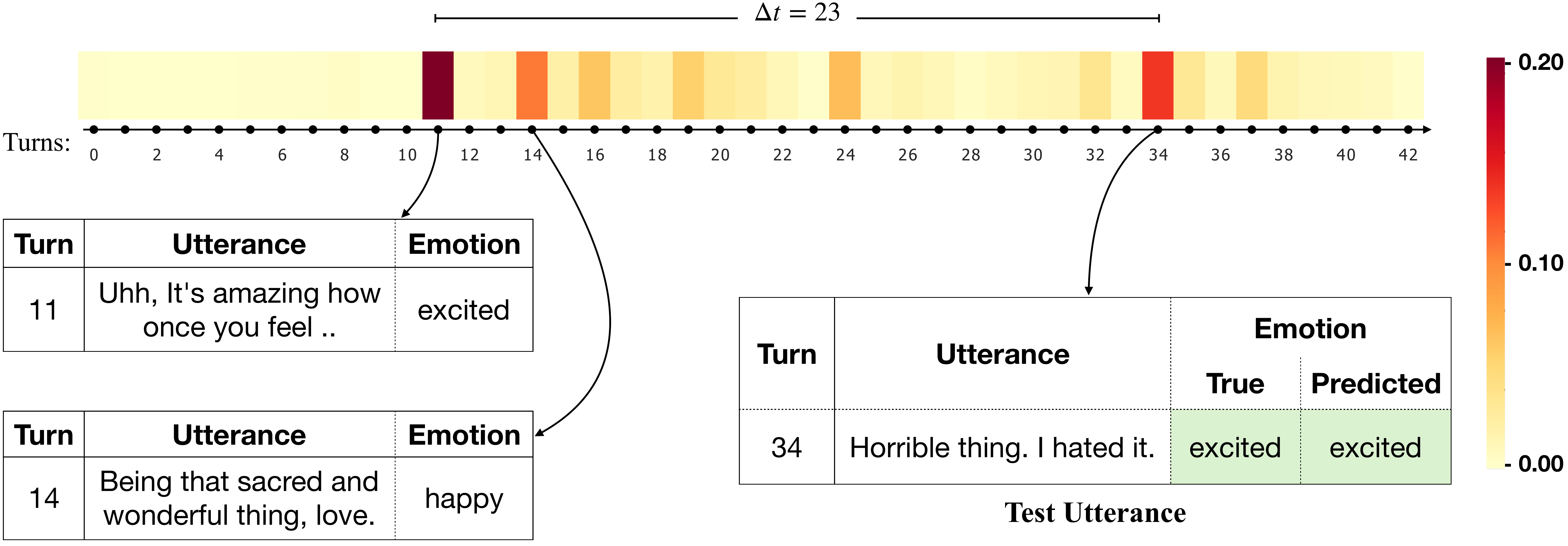} 
	\caption{}
	\label{fig:DistantAttention}
\end{subfigure}%\hspace{0.8cm}
  ~
\begin{subfigure}{0.49\textwidth}
	\centering
  \small
	\includegraphics[width=0.7\linewidth]{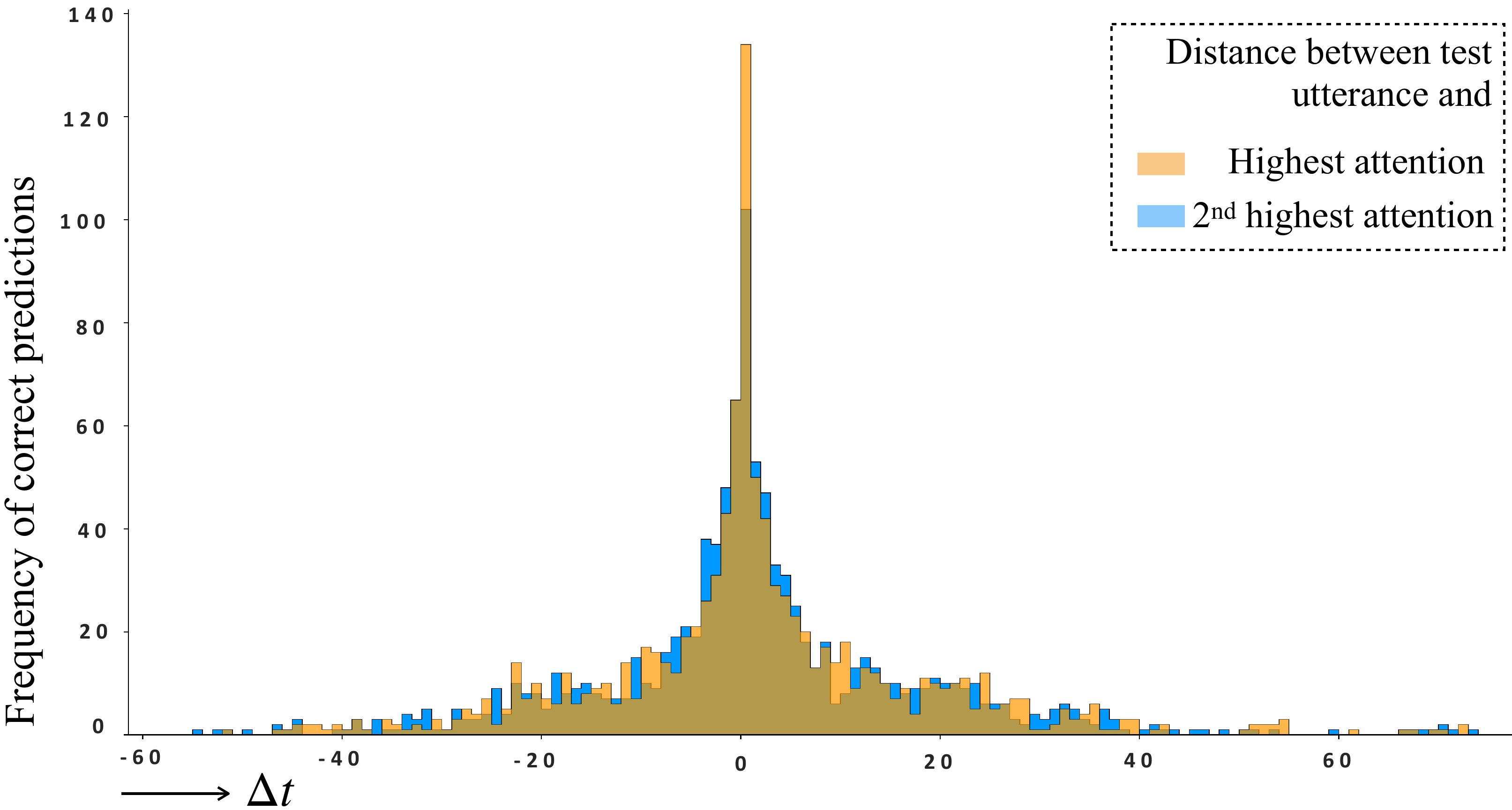} 
	\caption{}
	\label{fig:DeltaTTrend}
\end{subfigure}
\caption{(a) Illustration of the $\beta$ attention over emotion representations $e_t$;
  (b) Comparison of attention scores over utterance history of CMN and DialogueRNN ($\alpha$ attention). (c) An example of long-term dependency among utterances. (d) Histogram of $\Delta t=$ distance between the target utterance and its context utterance based on $\beta$ attention scores.}
\end{figure*}

\cref{fig:case-study} visualizes the $\beta$ (\cref{eq:beta}) attention over the emotion
representations $e_t$ for a segment of a conversation between a couple.
In the discussion, the woman ($P_A$) is initially at a neutral state, whereas the man ($P_B$) is angry throughout.
The figure reveals that the emotional attention of the woman is localized to the duration of her neutral state
(turns 1-16 approximately). For example, in the dialogue, turns $5, 7$, and $8$ strongly attend to turn $8$. Interestingly, turn $5$ attends to both past (turn $3$)
and future (turn $8$) utterances. Similar trend across other utterances establish inter-dependence between emotional states of future and past utterances. %\hl{Visualizing attentions also reveal that at any given time, speaker's emotion
%may depend on both speaker's and other party's previous turns.} As shown in \cref{fig:case-study}, emotion of turns $5$ and $7$ depend on respectively $3$ and $6$ which are turns of the other party. Such inter-party dependency is very much visible when at the time of emotion shift of one party (Figure XXX).
The beneficial consideration of future utterances through $GRU_{\mathcal{E}}$ is also apparent through turns $6, 9$. These utterances focus on the distant future (turn $49, 50$) where the man is at an enraged state, thus capturing emotional correlations across time. Although, turn $6$ is misclassified by our model, it still manages to infer a related emotional state (\textit{anger}) against the correct state (\textit{frustrated}). We analyze more of this trend in section~\ref{sec:error-analysis}.

\subsubsection{Dependency on distant context} 
%\begin{figure}[b] 
%	\centering 
%  \small
%	\includegraphics[width=0.8\linewidth]{delta_t_trend} 
%	\caption{Histogram of $\Delta t=$ distance between the target utterance and its context utterance based on
%	$\beta$ attention scores.}
%	\label{fig:DeltaTTrend}
%\end{figure}

%\begin{figure}[t] 
%	\centering 
%  \small
%	\includegraphics[width=0.8\linewidth]{distant_attention} 
%	\caption{An example of long-term dependency among utterances.}
%	\label{fig:DistantAttention}
%\end{figure}

For all correct predictions in the IEMOCAP test set in~\cref{fig:DeltaTTrend} we summarize the distribution over the relative distance between test utterance and ($2^{nd}$) highest attended utterance -- either in the history or future -- in the conversation. This reveals a decreasing trend with the highest dependence being within the local context. However, a significant portion of the test utterances ($\sim18\%$), attend to utterances that are $20$ to $40$ turns away from themselves, which  highlights the important role of long-term emotional dependencies. Such cases primarily occur in conversations that maintain a specific affective tone and do not incur frequent emotional shifts.
\cref{fig:DistantAttention} demonstrates a case of long-term context dependency. The presented conversation maintains a \emph{happy} mood throughout the dialogue. Although the $34^{th}$ turn comprising the
sentence \emph{Horrible thing. I hated it.}
seems to be a negative expression, when seen with the global
context, it reveals the \emph{excitement} present in the speaker. To disambiguate such cases, our model attends
to distant utterances in the past (turn $11$, $14$) which serve as prototypes of the emotional tonality of the
overall conversation. 

\subsection{Error Analysis}
\label{sec:error-analysis}
A noticeable trend in the predictions
is the high level of cross-predictions amongst related emotions. Most of the misclassifications by the model for \emph{happy} emotion are for \emph{excited} class. Also, \emph{anger} and \emph{frustrated} share misclassifications amongst each other. We suspect this is due to subtle difference between those emotion pairs, resulting in harder disambiguation.
Another class with high rate of false-positives is the \emph{neutral} class. Primary reason for this could be its majority in the class distribution over the considered emotions.

At the dialogue level, we observe that a significant amount of errors occur at turns having a change of
emotion from the previous turn of the same party. Across all the
occurrences of these \emph{emotional-shifts} in the testing set, our model correctly predicts $47.5\%$ instances. This stands less as compared to the $69.2\%$ success that it achieves at regions of no
\emph{emotional-shift}. Changes in emotions in a dialogue is a complex phenomenon governed by latent dynamics. Further improvement of these cases remain as an open area of research.

\subsection{Ablation Study}
\label{sec:ablation-study}

The main novelty of our method is the introduction of party state and emotion
GRU ($GRU_{\mathcal{E}}$). To comprehensively study the impact of these two components, we
remove them one at a time and evaluate their impact on IEMOCAP.
\begin{table}[t]
  \centering
  %\small
  \begin{tabular}{|c|c|c|}
  \hline
  \begin{tabular}{c}
     Party State
  \end{tabular} & \begin{tabular}{c}
  Emotion GRU
  \end{tabular} & F1 \\
  \hline
     -&+&55.56\\
     +&-&57.38\\
     +&+&59.89\\
     \hline
  \end{tabular}
  \caption{Ablated DialogueRNN for IEMOCAP dataset.}
  \label{tab:ablation}
\end{table}

As expected, following \cref{tab:ablation}, party state stands very important, as without its presence
the performance falls by $4.33\%$. We suspect that party state helps in extracting useful contextual information relevant to parties' emotion. %We suspect that his is due to party state being an indication of the
%parties' cognitive state, which is strongly related to their emotion.

Emotion GRU is also impactful, but less than party state, as its absence causes performance to fall by only $2.51\%$.
We believe the reason to be the lack of context flow from the other parties' states through the emotion representation of the preceding utterances.

%In conclusion, party state is more relevant to the emotion of an utterance than the emotion of the preceding
%utterances.

\section{Conclusion}
\label{sec:conclusion}

%We have presented an RNN-based neural architecture for emotion detection 
%in a conversation.
%In contrast to the state-of-the-art method, CMN, our method models each party in a more effective way.
%In contrast to the state-of-the-art method, CMN, our method treats each incoming utterance based on its speaker,
%which gives finer context to
%the utterance. 
We have presented an RNN-based neural architecture for emotion detection in a conversation. In contrast to the state-of-the-art method, CMN, our method treats each incoming utterance taking into account the characteristics of the speaker, which gives finer context to the utterance.
%Leveraging this idea, 
Our model outperforms the current state-of-the-art on two distinct datasets in both textual and multimodal settings. Our method is designed to be scalable to multi-party setting with more than two speakers, which we plan to explore in future work. %though we could not test it due to unavailability of a multi-party conversation dataset with emotion labels. This is left to our future work. %which is a part of our future work. 
%In our future work, we plan to explore this.

\bibliographystyle{aaai.bst}
\bibliography{bibexport_new.bib}

\end{document}